\documentclass[10pt,twocolumn,letterpaper]{article}

\usepackage[accsupp]{axessibility} % Improves PDF readability for those with disabilities.
\usepackage{iccv}
\usepackage{times}
\usepackage{epsfig}
\usepackage{graphicx}
\usepackage{amsmath}
\usepackage{amssymb}

\usepackage{subfigure}
\usepackage{overpic}
\usepackage[table,xcdraw]{xcolor}
\usepackage{enumitem}
\usepackage{threeparttable}
\usepackage{float}

\usepackage[breaklinks=true,bookmarks=false]{hyperref}

\iccvfinalcopy % *** Uncomment this line for the final submission

\def\iccvPaperID{1731} % *** Enter the ICCV Paper ID here

\ificcvfinal\fi

\begin{document}

%%%%%%%%% TITLE
\title{Regularizing Nighttime Weirdness: Efficient Self-supervised Monocular Depth Estimation in the Dark}

\author{Kun Wang$^{1}$\thanks{Contributes equally}\ , Zhenyu Zhang$^{2,1}$\footnotemark[1]\ , Zhiqiang Yan$^1$, Xiang Li$^1$, Baobei Xu$^3$, Jun Li$^{1}$\thanks{Corresponding authors}\ \ and Jian Yang$^{1}$\footnotemark[2]\\
\\
$^1$PCA Lab\thanks{PCA Lab, Key Lab of Intelligent Perception and Systems for High-Dimensional Information of Ministry of Education, and Jiangsu Key Lab of Image and Video Understanding for Social Security, School of Computer Science and Engineering, Nanjing University of Sci \& Tech.}\ , Nanjing University of Science and Technology, China\\
$^2$Tencent YouTu Lab \ \ \ \ \ \ \ \ \ \ \ $^3$Hikvision Research Institute\\

{\tt\small \{kunwang, yanzq, xiang.li.implus, junli, csjyang\}@njust.edu.cn}\\
{\tt\small zhangjesse@foxmail.com, 21625177@zju.edu.cn}
}

\maketitle
\ificcvfinal\thispagestyle{empty}\fi

\newcommand{\junli}[1]{{#1}}
\newcommand{\implus}[1]{\textcolor{cyan}{#1}}
\newcommand{\implusrewrite}[1]{{\textcolor{red}{#1}}}
\newcommand{\baobei}[1]{\textcolor{green}{#1}}
\newcommand{\todo}[1]{\textcolor{yellow}{#1}}
\newcommand{\alt}[1]{\textcolor{red}{(#1)}}

%%%%%%%%% ABSTRACT
\begin{abstract}
    Monocular depth estimation aims at predicting depth from a single image or video. Recently, self-supervised methods draw much attention since they are free of depth annotations and achieve impressive performance on several daytime benchmarks. However, they produce weird outputs in more challenging nighttime scenarios because of low visibility and varying illuminations, which bring weak textures and break brightness-consistency assumption, respectively. To address these problems, in this paper we propose a novel framework with several improvements: (1) we introduce Priors-Based Regularization to learn distribution knowledge from unpaired depth maps and prevent model from being incorrectly trained; (2) we leverage Mapping-Consistent Image Enhancement module to enhance image visibility and contrast while maintaining brightness consistency; and (3) we present Statistics-Based Mask strategy to tune the number of removed pixels within textureless regions, using dynamic statistics. Experimental results demonstrate the effectiveness of each component. Meanwhile, our framework achieves remarkable improvements and state-of-the-art results on two nighttime datasets. Code is available at \url{https://github.com/w2kun/RNW}.
\end{abstract}

%%%%%%%%% BODY TEXT
\vspace{-0.35cm}
\section{Introduction}
Monocular depth estimation is a fundamental topic in computer vision as it has wide range of applications in augmented reality \cite{intro_ar}, robotics \cite{intro_robotics} and autonomous driving \cite{intro_auto}, \etc. \junli{It often needs dense depth maps to learn the mapping from color images to the depth maps in supervised settings \cite{dorn, vnl, pap}. However, high-quality depth data are costly collected in a broad range of environments by using expensive depth sensors (\eg LiDAR and TOF).} \junli{Hence, many} efforts have been made to develop self-supervised approaches \cite{godard_stereo, zhou_method, selfattention2020, qinghuamethod}, which train a depth network to estimate depth maps by exploring geometry cues \junli{in videos, \ie, reconstructing a target view (or frame) from another view, instead of utilizing high-quality depth data. Furthermore, their performances are comparable to the supervised methods in well-lit environments, such as KITTI \cite{kitti} and Cityscapes \cite{cityscapes}. Unfortunately, there are a very few works to handle with more challenging nighttime scenarios. Thus we focus on nighttime self-supervised depth estimation.}
%However, when the self-supervised methods are directly applied in more challenging nighttime scenarios, they fail miserably and produce weird outputs (see Fig.\ref{fig.1 mono2}).
%due to low visibility and varying illuminations (see Fig.\ref{fig.1 color})
\begin{figure}
    \begin{center}
        \subfigure{
            \includegraphics[width=0.95\linewidth]{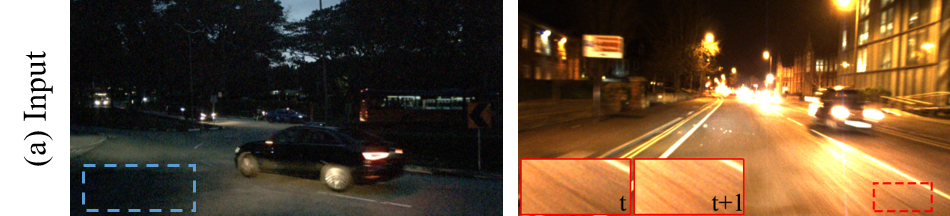}
            \label{fig.1 color}
        }
        
        \vspace{-0.5cm}
        \subfigure{
            \includegraphics[width=0.95\linewidth]{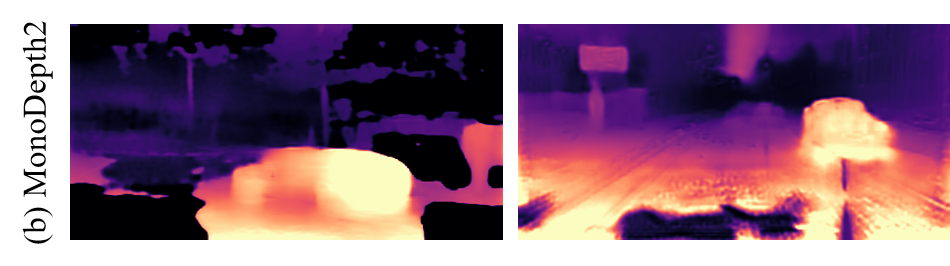}
            \label{fig.1 mono2}
        }
        
        \vspace{-0.5cm}
        \subfigure{
            \includegraphics[width=0.95\linewidth]{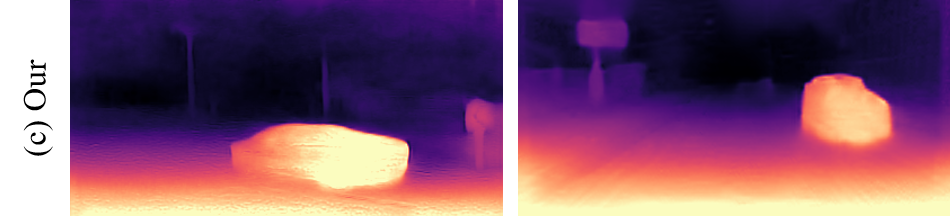}
            \label{fig.1 our}
        }
    \end{center}
    \vspace{-0.3cm}
    \caption{\junli{Depth from nuScenes (left) and RobotCar (right). (a) Input images: \textcolor{cyan}{cyan dashed} box indicates an textureless patch caused by the low visibility (\eg, dark), and two \textcolor{red}{red} borders illustrate the varying lights between $t$ and $t+1$ frames. (b) shows that low visibility and varying lights result in big holes and non-smoothness in the depth maps using MonoDepth2 \cite{monodepth2}, respectively. (c) demonstrates depth predictions of our framework. }}
    \label{fig.1}
    \vspace{-0.5cm}
\end{figure}

\junli{Actually, the nighttime scenario includes two important problems, low visibility and varying illuminations, resulting in that most of the existing self-supervised methods (\eg, MonoDepth2 \cite{monodepth2}) produce a weird depth output (see Fig.\ref{fig.1 mono2}).}  1) \emph{Low visibility usually creates textureless areas.} \junli{For example, the cyan dashed box in the left of Fig.\ref{fig.1 color} shows a dark region with indistinguishable visual texture.} This textureless aggravates \junli{depth maps with big holes in the left of Fig. \ref{fig.1 mono2}, though they may be used to correctly reconstruct the target view by sampling nearby pixels with similar brightness.} 2) \emph{Varying illuminations from flickering streetlights or moving cars, undermine the brightness consistency assumption in the right of Fig.\ref{fig.1 color},} where the two image patches with different brightness are cropped from the same place of two temporally adjacent frames. \junli{This inconsistency brings an imperfect reconstruction of the target view, that is, a high training loss, which also produces non-smooth depth map in the right of Fig. \ref{fig.1 mono2}. Clearly, the incorrect depth prediction (\eg, non-smoothness and big holes) indicates a failure of training the depth network.}
%an imperfect reconstruction of the target view, that is, a high training loss, and 
%even with correct depth output, target view may not be perfectly reconstructed
%Our method is more stable in the dark, owing to several improvements.

\junli{To address these two problems, in this paper we propose an efficient nighttime self-supervised framework for depth estimation with three improvements. Firstly, we introduce Priors-Based Regularization (PBR) module to constrain the incorrect depth in neighborhoods of depth references, and prevent the depth network from being incorrectly trained. This constraint is implemented by learning prior depth distribution from unpaired references in an adversarial manner. Furthermore, 2D coordinates are encoded as an additional input of PBR to find useful depth distribution, which is related with its pixel location.
    Secondly, we leverage Mapping-Consistent Image Enhancement (MCIE) module to deal with the low visibility. Although image enhancement methods, \eg, Contrast Limited Histogram Equalization (CLHE) \cite{clahe}, can be used to achieve remarkable results on low-light images \cite{RetinexNet, enlightengan}, they are difficult to handle the correspondence among video frames, which is essential to self-supervised depth estimation. Thus, we extend the CLHE method to keep brightness consistency while enhancing low-visible video frames. 
    %Although image enhancement is an efficient way for this problem, previous methods \cite{RetinexNet, enlightengan} usually lack brightness correspondence among frames. In contrast, MCIE executes identical mapping function to each frame and ensures brightness consistency after enhancement.
    % ADD
    Finally, we present Statistics-Based Mask (SBM) to tackle textureless regions. Though Auto-Mask \cite{monodepth2} is a widely used strategy to efficiently choose textureless regions, its dependence on photometric loss makes it unable to adjust the numbers of removed pixels. To compensate this weakness, we introduce SBM to better handle nighttime scenarios by flexibly tuning masked pixels using dynamic statistics.} In short, our contributions can be summarized as three-fold: 
\begin{itemize}[leftmargin=*]
    \setlength{\itemsep}{0pt}
    \item  We \junli{propose} Priors-Based Regularization module to learn distribution knowledge from unpaired references and prevent model from being incorrectly \junli{trained}.
    \item We leverage Mapping-Consistent Image Enhancement module to deal with low visibility in the dark and maintain brightness consistency.
    \item We \junli{present} Statistics-Based Mask to better handle textureless regions, by using dynamic information. Together, these contributions yield state-of-the-art performance in nighttime depth estimation task and efficiently reduce the weirdness in depth outputs.
\end{itemize}

%------------------------------------------------------------------------
\section{Related Work}

\noindent\textbf{Self-supervised Depth Learning from Videos}. SfMLearner \cite{zhou_method} is a pioneering work in this task. It jointly learns to predict depth and relative pose of the camera, which is supervised by reconstruction of target frame. This process is based on the assumption of static scene while moving objects violate it. To address this problem, previous works have employed optical flow \cite{dfnet2018, geonet, competitive} and pre-trained segmentation models \cite{inthewild, signet2019, casser2019depth} to compensate and mask pixels within moving objects, respectively. Occlusion is also a challenge. MonoDepth2 has provided a minimum reprojection loss 
to deal with it. Besides, approaches with geometry priors, such as normal \cite{depthnormal, occlusion2020} and geometry consistency \cite{bian} have been exploited for better performance. Recently, PackNet \cite{3dpack} has proposed a novel network architecture to learn detail-preserving representations. FM \cite{fm} have leveraged more informative feature metric loss to address the problem of textureless regions. These methods offer ideas to improve the performance of self-supervised depth estimation in daytime environments but think little of more challenging nighttime scenarios.
\\
\noindent\textbf{Nighttime Self-supervised Learning Methods}. Nighttime self-supervised depth estimation is a relatively under-explored topic as a result of its numerous challenges. Existing works have explored approaches to predict depth from thermal images \cite{multispectral2018, thermal-night}. However, thermal images have less texture details and limited resolution. Thermal cameras are also expensive. Defeat-Net \cite{defeat-net} has been proposed to simultaneously learn cross-domain feature representation and depth estimation to acquire a more robust supervision. Nevertheless, it is unable to tackle the low visibility and varying illuminations. ADFA \cite{adfa} considers this problem as one of domain adaption and has adapted a network trained on daytime data to work for nighttime images. It aims to transfer knowledge from daytime to nighttime data.
Different from ADFA, we only use prior depth distribution from daytime data as regularization and directly exploit depth estimation knowledge from nighttime scenes.
\begin{figure*}[t]
    \begin{center}
        \includegraphics[width=0.92\linewidth]{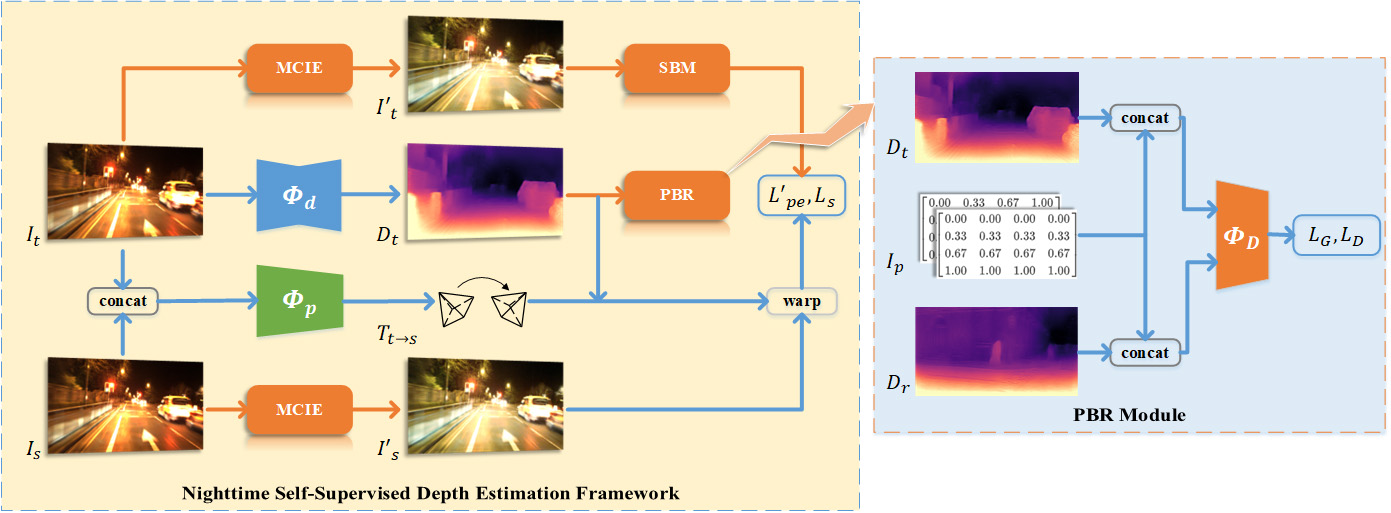}
    \end{center}
    \vspace{-0.2cm}
    \caption{Overall pipeline with three proposed improvements ({\color{orange}orange} boxes): Priors-Based Regularization (PBR), Mapping-Consistent Image Enhancement (MCIE) and Statistics-Based Mask (SBM). PBR is shown on the right. The \emph{concat} means concatenation operation along channel dimension, $D_r$ and $I_p$ separately denotes the referenced depth map and coordinates image.}
    \label{fig.overall}
    \vspace{-0.3cm}
\end{figure*}
\\
\noindent\textbf{Domain Adaptation in Depth Estimation}. Domain adaptation (DA) \cite{adaptationsurvey2018}, as a subproject of transfer learning \cite{transfersuervey2020}, aims to efficiently leverage prior knowledge learned from source domain. In depth estimation, an important application is to close the gap \cite{atapour2018real, t2net, gu2020coupled} between synthetic \cite{vkitti2, sintel} and real-world data, for mitigating the need for large-scale real-world ground truth. For better use of geometry structure, GASDA \cite{gasda} has been performed to exploit epipolar geometry of stereo images. CoMoDA \cite{comoda} has been applied to continuously adapt a pre-trained model on test videos. The prior knowledge is also employed in our framework but is to regularize training.
\\
\noindent\textbf{Low-light Image Enhancement}. Image enhancement is a resultful approach to improve the brightness and contrast of images. Retinex \cite{retinex} decomposes an image into reflectance and illumination. Histogram equalization based methods (\eg CLHE \cite{clahe}) readjustment the brightness level of pixels. Recently, methods \cite{RetinexNet, kindlingthedarkness} combining Retinex with Convolutional Neural Network (CNN) have shown impressive results. Previous learning-based works \cite{multi-exposure, seeinthedark} require paired data. To address this problem, efforts have been put into exploring approaches with unpaired inputs \cite{enlightengan} or zero-reference \cite{zeroref}. Although these methods have proven to be effective, they pay no attention to the brightness correspondence among frames which is essential for self-supervised training of depth estimation.
%---------------------------------------------------------------------------------------------------------------------
\section{Method}

In this section, we propose a novel self-supervised framework to learn depth estimation from nighttime videos. Before presenting it, we first introduce a basic self-supervised training with necessary notations. %for depth estimation.

\subsection{Self-supervised Training}
In self-supervised depth estimation, the learning problem is considered as a view-synthesis process. It reconstructs target frame $I_t$ from the viewpoint of each source image $I_s$ by performing a reprojection using depth $D_t$ and relative pose $T_{t\rightarrow s}$. In the setting of monocular training, $D_t$ and $T_{t\rightarrow s}$ are predicted by two neural networks via $D_t=\Phi_d(I_t)$ and $T_{t\rightarrow s}=\Phi_p(I_t,I_s)$, respectively. The camera intrinsic parameter $K$ is also required for projection operations. With the above variables, we can acquire a per-pixel correspondence between an arbitrary point $p_t$ in $I_t$ and another point $p_s$ in $I_s$ by
\begin{equation}
    p_s\sim KT_{t\rightarrow s}D_t(p_t)K^{-1}p_t,
\end{equation}
After that, $I_t$ can be reconstructed from $I_s$ with differentiable bilinear sampling \cite{spatialtransformer} operation $s(\cdot,\cdot)$:
\begin{equation}
    \hat{I_t}=s(I_s,p_s).
\end{equation}
The model learning is based on the above warping process, \ie reconstruct target frame from source view, and the objective is to reduce reconstruction error by optimizing $\Phi_d$ and $\Phi_p$ to produce more accurate outputs. Following \cite{godard_stereo}, we apply $\ell_1$ and SSIM \cite{ssim} together as photometric error to measure the difference between $I_t$ and $\hat{I_t}$,
\begin{equation}
    \begin{aligned}
        L_{pe}(I_t,\hat{I_t})=\frac{\alpha}{2}(1-SSIM(I_t,\hat{I_t}))+\\(1-\alpha)\|I_t-\hat{I_t}\|_1,
    \end{aligned}
\end{equation}
where $\alpha$ is set to 0.85 in all experiments.

Moreover, this is an ill-posed problem as there are a large amount of possible incorrect depths which lead to the correct reconstruction of target frame given the relative pose $T_{t\rightarrow s}$. To address this depth ambiguity, we follow previous works \cite{godard_stereo} by applying edge-aware smoothness loss to enforce smoothness in depths,
\begin{equation}\label{eq.Ls}
    L_s=|\partial_x D_t|e^{-|\partial_x I_t|}+|\partial_y D_t|e^{-|\partial_y I_t|},
\end{equation} 
where $\partial_x$ and $\partial_y$ are image gradient along horizontal and vertical axes, respectively.

\subsection{Nighttime Depth Estimation Framework}
Here, we present the nighttime self-supervised depth estimation framework, \junli{which is illustrated in Fig. \ref{fig.overall}}. The framework contains three improvements for nighttime environments, including PBR, MCIE and SBM, which are detailedly described next.

\subsubsection{Priors-Based Regularization}
\junli{Priors-Based Regularization (PBR) is to constrain the depth output in neighborhoods of depth references using adversarial manner, which is shown on the right of Fig. \ref{fig.overall}. The depth estimation network $\Phi_d$ is considered as a generator and a discriminator $\Phi_D$ using Patch-GAN \cite{patchgan} is employed in PBR. Adversarial depth maps are $(D_t, D_r)$, where depth output $D_t$ is generated by $\Phi_d$, and $D_r$ is a referenced depth map. The discriminator is used to distinguish $D_t$ and $D_r$, while $\Phi_d$ tries to make its output indistinguishable with $D_r$.} In order to acquire the referenced depth maps, we train a depth estimation network $\Phi'_d$ to produce $D_r$ in a self-supervised manner using a daytime dataset. Note that, $D_t$ and $D_r$ are unpaired, thus the same scene as nighttime dataset is not required.

Besides, we find a close relationship between depth of a pixel and its position. For example, an image of driving scene usually \junli{shows} a view from road to sky along vertical direction. Based on this observation, we encode the 2D coordinates of each pixel into an image $I_p$ as an additional input of $\Phi_D$. $I_p$ is composed of two single-channel maps separately indicating the coordinates along $x$ and $y$ axes and is scaled to range $[0, 1]$ for normalization. Furthermore, both $D_t$ and $D_r$ are scale-ambiguity, hence it is unreasonable to unify their scales. We apply $\mu(\cdot)$ to perform normalization in depth maps to address the misalign of their scales,
\begin{equation}
    \mu(D) = D / avg(D),
\end{equation}
where $avg(D)$ computes average along space dimension. Let $cat(\cdot, \cdot)$ denote the concatenation operation along channel dimension, $\omega_d$ and $\omega_D$ are network weights of $\Phi_d$ and $\Phi_D$, $\{D_t\}$ and $\{D_r\}$ present a set of $D_t$ and $D_r$, respectively, then the optimization objective for PBR can be written as
\begin{equation}\label{eq.lsgan_loss}
    \begin{aligned}
        \min_{\omega_D}L_D=&\frac{1}{2}\mathbb{E}_{D_r\in \{D_r\}}[(\Phi_D(cat(I_p, \mu(D_r))) - 1)^2]+\\ &\frac{1}{2}\mathbb{E}_{D_t\in \{D_t\}}[\Phi_D(cat(I_p, \mu(D_t)))^2]\\
        \min_{\omega_d}L_G=&\frac{1}{2}\mathbb{E}_{D_t\in \{D_t\}}[(\Phi_D(cat(I_p, \mu(D_t)))-1)^2],
    \end{aligned}
\end{equation}
in which the loss format in \cite{lsgan} is adopted for better convergence.

\textbf{Remark}. It is hard to instantiate the depth of a specific sample from a general depth distribution, since the depths are distributed in a range rather than a certain value. But it is easier to find a outlier (in our case weird depth value) as it greatly deviate from expected outputs. This is the reason why we use PBR as a regularizer. In addition, the application for PBR is not limited to nighttime depth estimation, but can also be extended to other similar tasks.

\subsubsection{Mapping-Consistent Image Enhancement}
Mapping-Consistent Image Enhancement (MCIE) is adapted from Contrast Limited Histogram Equalization\cite{clahe} (CLHE) algorithm to meet the need for keeping brightness-consistency, which is essential for self-supervised depth estimation. This is implemented by using a brightness mapping function $b'=\gamma(b)$ and applying it to target frame and source frames together, \ie
\begin{equation}
    I'_t=\gamma(I_t),I'_s=\gamma(I_s).
\end{equation}
$\gamma$ is a single-value mapping function, which maps an input brightness to single certain output. By this way, the brightness consistency among target and source frames is naturally maintained.

\begin{figure}
    \begin{center}
        \includegraphics[width=0.9\linewidth]{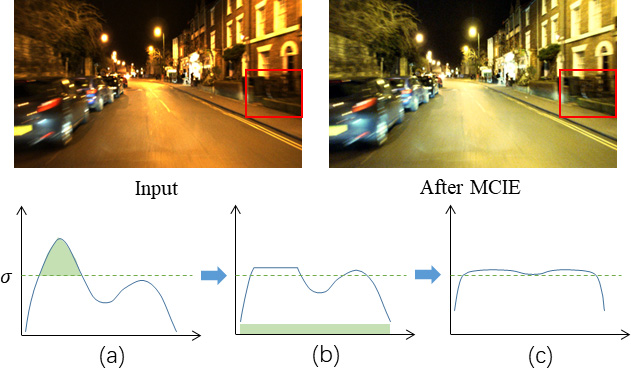}
    \end{center}
    \vspace{-0.2cm}
    \caption{The top two images illustrate the effectiveness of MCIE, in which we can see an obvious improvement on visibility, especially within the red box. The bottom three figures show the main steps to compute the brightness mapping function $\gamma$.}
    \label{fig.clhe}
    \vspace{-0.2cm}
\end{figure}

We show the primary steps to compute $\gamma$ at the bottom of Fig. \ref{fig.clhe}. Supposing the frequency distribution $f_b=h(b)$ of input image is given, where $f_b$ is the frequency of brightness level $b$.
%Before that, the frequency distribution $f_b=h(b)$ of a given image need to be calculated, where $f_b$ is the frequency of brightness level $b$. 
Firstly, we clip the frequency greater than the preset parameter $\sigma$ to avoid the amplification of noise signal. Secondly, the clipped frequency is evenly filled to each brightness level, as shown in the subfigure (b). Finally, $\gamma$ can be obtained with cumulative distribution $cdf$ through
\begin{equation}
    \gamma(b)=\frac{cdf(b)-cdf_{min}}{cdf_{max}-cdf_{min}}\times (L-1),
\end{equation}
where $cdf_{min}$ and $cdf_{max}$ separately indicate the minimum and maximum of $cdf$, $L$ presents the number of brightness level (commonly 256 in color images).

MCIE brings higher visibility and more details to nighttime images. We illustrate it with the top two images in Fig. \ref{fig.clhe}, where we can see a remarkable improvement on brightness and contrast, especially within the area framed by red box. MCIE only enhances image when computing photometric loss and doesn't change the input of networks. It redefine the warping process as
\begin{equation}
    \hat{I'_t}=s(\gamma(I_s),p_s).
\end{equation}
Accordingly, the photometric loss is adapted to use enhanced images through
\begin{equation}\label{eq.Lpe}
    \begin{aligned}
        L'_{pe}(I'_t,\hat{I'_t})=\frac{\alpha}{2}(1-SSIM(I'_t,\hat{I'_t}))+\\(1-\alpha)\|I'_t-\hat{I'_t}\|_1.
    \end{aligned}
\end{equation}

\subsubsection{Statistics-Based Mask}
We introduce Statistics-Based Mask (SBM) to compensate Auto-Mask \cite{monodepth2} (AM) strategy as it is unable to adjust the number of removed pixels due to the dependence on photometric loss. Let $[\ ]$ denote Iverson bracket. AM produces a mask between target and source frames by
\begin{equation}
    m_{a}=[L_{pe}(I_t,\hat{I_t})<L_{pe}(I_t,I_s)].
\end{equation}
Unlike AM, SBM uses dynamic statistics to flexibly tune the masked pixels. During training, SBM computes the difference between target frame and each source frame by $d_{ts}=\|I_t-I_s\|_1$ and employs Exponentially Weighted Moving Average (EWMA) to obtain the mean $d_{ts}$ in recent samples, which is figured by
\begin{equation}
    \tilde{d_{ts}}(i)=\beta\times \tilde{d_{ts}}(i-1) + (1-\beta)\times d_{ts}(i),
\end{equation}
where $i$ is the current time and $\beta$ is momentum parameter that is set to 0.98 in our experiments. It is more stable and reflects global statistics. To tune masked pixels, a parameter $\epsilon\in[0,100]$ indicating the percentile of $\tilde{d_{ts}}$ requires to be defined and can be used to generate the mask $m_s\in\{0,1\}$ between target and source frames through
\begin{equation}
    m_s=[d_{ts}>p(\tilde{d_{ts}}, \epsilon)],
\end{equation}
where $p(\tilde{d_{ts}}, \epsilon)$ computes the $\epsilon$th percentile of $\tilde{d_{ts}}$. We combine $m_a$ with $m_s$ via element-wise product to make the final mask used in our framework, \ie
\begin{equation}
    m=m_a\odot m_s.
\end{equation}

\begin{figure}
    \begin{center}
        \subfigure[Input]{
            \includegraphics[width=0.32\linewidth]{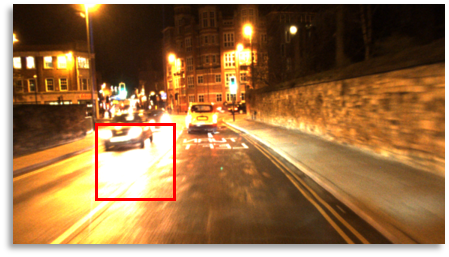}
        }
        \hspace{-0.36cm}
        \subfigure[$m_s$]{
            \includegraphics[width=0.32\linewidth]{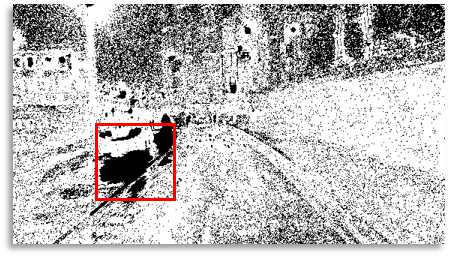}
        }
        \hspace{-0.36cm}
        \subfigure[$m_a$]{
            \includegraphics[width=0.32\linewidth]{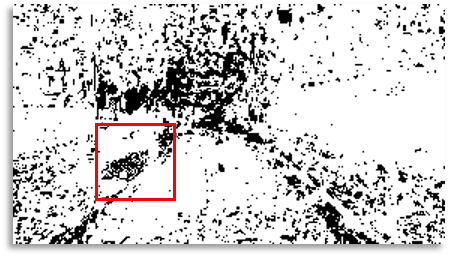}
        }
    \end{center}
    \vspace{-0.2cm}
    \caption{Visual comparison between $m_s$ and $m_a$\cite{monodepth2}, where black pixels are removed from loss. We can see that, $m_s$ can better mask textureless regions (\eg light spot in {\color{red}red} box).}
    \label{fig.mask}
    \vspace{-0.2cm}
\end{figure}

The visual comparison between $m_s$ and $m_a$ is shown in Fig. \ref{fig.mask}. It can be seen that $m_s$ is more effective in masking textureless regions (\eg the sky and the bright light spot framed by red box). We employ $m_s$ together with $m_a$, since $m_a$ can prevent large errors from being incorporated, which works like a regularizer.

\subsubsection{Final Loss}
In summary, the final loss is composed of photometric loss (Eq. \eqref{eq.Lpe}), edge-aware smoothness (Eq. \eqref{eq.Ls}) loss and PBR regularization (Eq. \eqref{eq.lsgan_loss}), \ie
\begin{equation}
    Loss=mL'_{pe} + \eta L_s + \xi L_G + \tau L_D,
\end{equation}
where $\eta$, $\xi$ and $\tau$ are weight parameters.
%------------------------------------------------------------------------
\section{Experiment}
\begin{table*}
    \begin{minipage}{0.75\linewidth}
        \resizebox{\linewidth}{!}{
            \footnotesize
            \renewcommand\arraystretch{1.2}
            \centering
            \begin{tabular}{c||c|c|c|c|c|c|c}
                \hline
                Method     & \cellcolor[RGB]{244,176,132}Abs Rel & \cellcolor[RGB]{244,176,132}Sq Rel & \cellcolor[RGB]{244,176,132}RMSE & \cellcolor[RGB]{244,176,132}RMSE log & \cellcolor[RGB]{155,194,230}$\delta_1$ & \cellcolor[RGB]{155,194,230}$\delta_2$ & \cellcolor[RGB]{155,194,230}$\delta_3$ \\ \hline
                \multicolumn{8}{c}{\cellcolor[RGB]{198,224,180}RobotCar-Night}   \\
                \underline{MonoDepth2} \cite{monodepth2}    & 0.3999            & 7.4511        & 6.6416        & 0.4429        & 0.7444        & 0.8921        & 0.9280            \\
                SfMLearner \cite{zhou_method}   & 0.6754            & 15.4334       & 9.4324        & 0.6046        & 0.5465        & 0.8003        & 0.8733            \\
                SC-SfMLearner \cite{bian}       & 0.6029            & 16.0173       & 9.2453        & 0.5620        & 0.7185        & 0.8722        & 0.9091            \\
                PackNet \cite{3dpack}           & 0.2836            & 4.0257        & 5.3864        & 0.3351        & 0.7425        & 0.9143        & 0.9560            \\
                FM \cite{fm}                    & 0.3953            & 7.5579        & 6.7002        & 0.4391        & 0.7605        & 0.8943        & 0.9299            \\
                DeFeat-Net \cite{defeat-net}
                & 0.3929            & 4.8955        & 6.3429        & 0.4236        & 0.6256        & 0.8290        & 0.8992            \\
                \hline
                MonoDepth2 (Day)                   & 0.3211            & 1.8672        & 4.9818        & 0.3568        & 0.4446        & 0.7813        & 0.9353            \\
                FM (Day)                           & 0.2928            & 1.5380        & 4.5951        & 0.3337        & 0.4888        & 0.8054        & 0.9497            \\
                Reg Only                        & 0.5006            & 3.7608        & 6.6351        & 0.7518        & 0.2841        & 0.5643        & 0.8156            \\
                \hline
                Our             & \textbf{0.1205}   & \textbf{0.5204}   & \textbf{2.9015}   & \textbf{0.1633}   & \textbf{0.8794}   & \textbf{0.9688}   & \textbf{0.9896}   \\
                \hline
                \multicolumn{8}{c}{\cellcolor[RGB]{198,224,180}nuScenes-Night}   \\
                \underline{MonoDepth2} \cite{monodepth2}  & 1.1848            & 42.3059           & 21.6129           & 1.5699            & 0.1842            & 0.3598            & 0.5044           \\ 
                SfMLearner \cite{zhou_method}   & 0.6004            & 8.6346            & 15.4351           & 0.7522            & 0.2145            & 0.4166            & 0.5961            \\
                SC-SfMLearner \cite{bian}   & 1.0508            & 30.5865           & 19.6004           & 0.8854            & 0.1823            & 0.3673            & 0.5422            \\
                PackNet \cite{3dpack}         & 1.5675            & 61.5101           & 25.8318           & 1.3717            & 0.1387            & 0.2980            & 0.4313            \\
                FM \cite{fm}             & 1.1383            & 41.6166           & 20.8481           & 1.1483            & 0.2376            & 0.4252            & 0.5650            \\
                % \hline
                % MonoDepth2(Mix)         & 1.0701        & 38.3358       & 20.1165       & 1.1910        & 0.2692        & 0.4511        & 0.5862 \\
                % FM(Mix)                 & 0.9563        & 34.0522       & 18.7942       & 0.7983        & 0.3054        & 0.5071        & 0.8693 \\
                % \hline
                Our             & \textbf{0.3150}   & \textbf{3.7926}   & \textbf{9.6408}   & \textbf{0.4026}   & \textbf{0.5081}   & \textbf{0.7776}   & \textbf{0.8959}   \\
                \hline
            \end{tabular} 
        }
    \end{minipage} %
    \begin{minipage}{0.18\linewidth}
        \caption{Quantitative results. We compare our framework with previous state-of-the-art methods on both RobotCar-Night and nuScenes-Night datasets. Baseline method is \underline{underlined} and the best results in each category are in \textbf{bold}. DeFeat-Net is tested with a checkpoint trained on RobotCar-Season \cite{robotcarseason}. (Day) indicates that the model is trained on another daytime datasets. Reg Only uses PBR regularization as the only loss in training.}
        \label{tab.compare}
    \end{minipage}
\end{table*}

In this section, the proposed framework is evaluated through series experiments and is compared with state-of-the-art (SOTA) methods. Before reporting it, we firstly introduce the RobotCar-Night and nuScenes-Night datasets, on which all methods are tested, then describe the implementation details. Finally, we show the ablation study that demonstrate the effectiveness of PBR, MCIE and SBM.

\subsection{Dataset}
\noindent\textbf{RobotCar-Night}. Oxford RobotCar \cite{RobotCarDatasetIJRR} dataset contains a large amount of data collected from one route through central Oxford, and covers various weather and traffic conditions. We build RobotCar-Night using the left images of the front stereo-camera (Bumblebee XB3) data from the sequences captured on 2014-12-16-18-44-24 and images are cropped to $1152\times 672$ for excluding car-hood. The training set is formed from the first 5 splits, in which frames while camera stops moving are removed. The front LMS laser sensor data and INS data are used along with the official toolbox, to generate depth ground truth for testing. For more accurate evaluation, we manually pick up high-quality outputs. As a result, the RobotCar-Night dataset contains more than 19k training sequences and 411 test samples.
\\
\noindent\textbf{nuScenes-Night}. nuScenes \cite{nuscenes2019} is a large-scale dataset for autonomous driving, which is composed of 1000 diverse driving scenes in Boston and Singapore, where each scene is presented by a video of 20 second length. We firstly select 60 nighttime scenes in total. These scenes are more challenging than RobotCar, due to lower visibility and more complicated traffic conditions. Images are firstly cropped to $1536\times 768$. The front camera data from part of scenes are used to build training set and data from top LiDAR sensor in rest scenes are employed with officially released toolbox to generate depth ground truth. In summary, nuScenes-Night contains more than 10k training sequences and 500 test samples.

\subsection{Implementation Detail}
Our depth estimation network is based on U-Net \cite{unet} architecture, \ie an encoder-decoder with skip connections. The encoder is a ResNet-50 \cite{resnet}, with fully-connected layer removed and maxpooling replaced by a stride convolution. Depth decoder contains five $3\times 3$ convolutional layers and uses nearest interpolation for up-sampling. Sigmoid and Leaky Relu nonlinearities are separately employed at the output and elsewhere. Pose prediction network $\Phi_p$ is structured with ResNet-18, and outputs a vector of six element length for each sample. $\Phi_D$ in PBR is a Patch-GAN \cite{patchgan} based discriminator with three convolutional layers of $4\times 4$ kernel size.

In experiments on RobotCar-Night, the two parameters in MCIE and SBM are set to $\sigma=0.008$ and $\epsilon=10$, respectively. In final loss, $\eta=1e^{-3}$, $\xi=2.5e^{-4}$ and $\tau=2.5e^{-4}$ are set. The data captured on 2014-12-09-13-21-02 from Oxford RobotCar is employed to train the network $\Phi'_d$. For nuScenes-Night, $\sigma$ and $\epsilon$ is set to $0.004$ and $20$, respectively. $\eta=1e^{-3}$, $\xi=4e^{-4}$ and $\tau=4e^{-4}$ are configured for final loss. Other scenes containing daytime images in nuScenes are used to train $\Phi'_d$. Notice that, the scene for training $\Phi'_d$ is not limited to daytime. The reasons of using daytime dataset relies on that models are more easier to trained on daytime environments. For more information about scene selection and parameter setting, please refer to supplementary material (Supp).

Our models are implemented in PyTorch\cite{pytorch}, trained for 20 epochs on four RTX2080TI GPUs using Adam\cite{adam} optimizer, with $576\times320$ and $768\times384$ input resolution for RobotCar-Night and nuScenes-Night, respectively. The learning rate is initialized as $3e^{-5}$, linearly warmed up to $1e^{-4}$ after 500 iterations and halved at 15th epochs. We apply seven standard metrics for testing, including Abs Rel, Sq Rel, RMSE, RMSE log, $\delta_1$, $\delta_2$ and $\delta_3$. For more information about test metrics, please see Supp. During evaluation, we restrict the maximum depth to 40m and 60m for RobotCar-Night and nuScenes-Night datasets, respectively. Moreover, the scale between predicted depth and ground truth depth is aligned using a scale factor introduced by \cite{zhou_method}
\begin{equation}
    \hat{s}=median(D_{gt})/median(D_{pred}).
\end{equation}
The predicted depth is multiplied with $\hat{s}$ before evaluation, which called median scaling.

\subsection{Compare with SOTA Methods}

\begin{figure*}
    \begin{center}
        \includegraphics[width=0.95\linewidth]{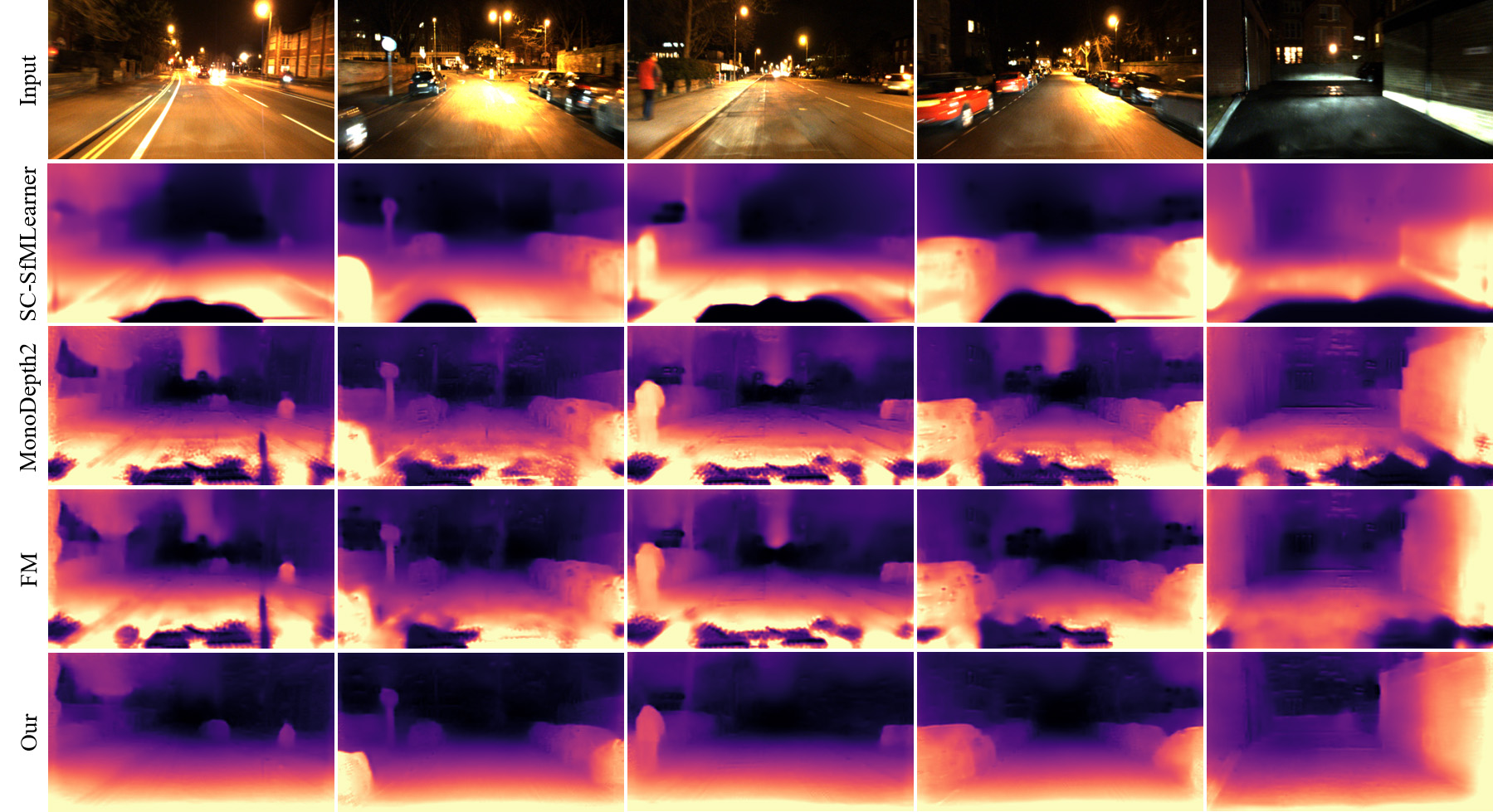}
    \end{center}
    \vspace{-0.2cm}
    \caption{Qualitative comparison on RobotCar-Night dataset. The top row is input image. Results from SC-SfMLearner \cite{bian}, MonoDepth2 \cite{monodepth2} and FM \cite{fm} are separately listed from the second to the forth row. Our results are shown at the bottom.}
    \label{fig.compare_rc}
    \vspace{-0.2cm}
\end{figure*}

Here, we compare \junli{our method} with several SOTA approaches, including SfMLearner \cite{zhou_method}, SC-SfMLearner \cite{bian}, MonoDepth2 \cite{monodepth2}, PackNet \cite{3dpack} and FM \cite{fm}. All methods are evaluated on both RobotCar-Night and nuScenes-Night datasets. The results are reported in Table \ref{tab.compare} and we choose the Sq Rel metric for subsequent analysis. In general, our method significantly outperforms other competitors and shows remarkable improvements on each evaluation metric. It improves the baseline method by $93.0\%$ and $91.0\%$ on RobotCar-Night and nuScenes-Night, respectively. Compared to PackNet, which employs expensive 3D convolution to learn detail-preserving representations, our method is more lightweight and separately achieves an improvement of $87.1\%$ and $93.8\%$ on the two datasets. Also, $93.1\%$ and $90.9\%$ improvements can be seen in comparison with the recent FM method, which introduces feature-metric loss to constrain the loss landscapes to form proper convergence basins.

\begin{figure}
    \begin{center}
        \includegraphics[width=0.95\linewidth]{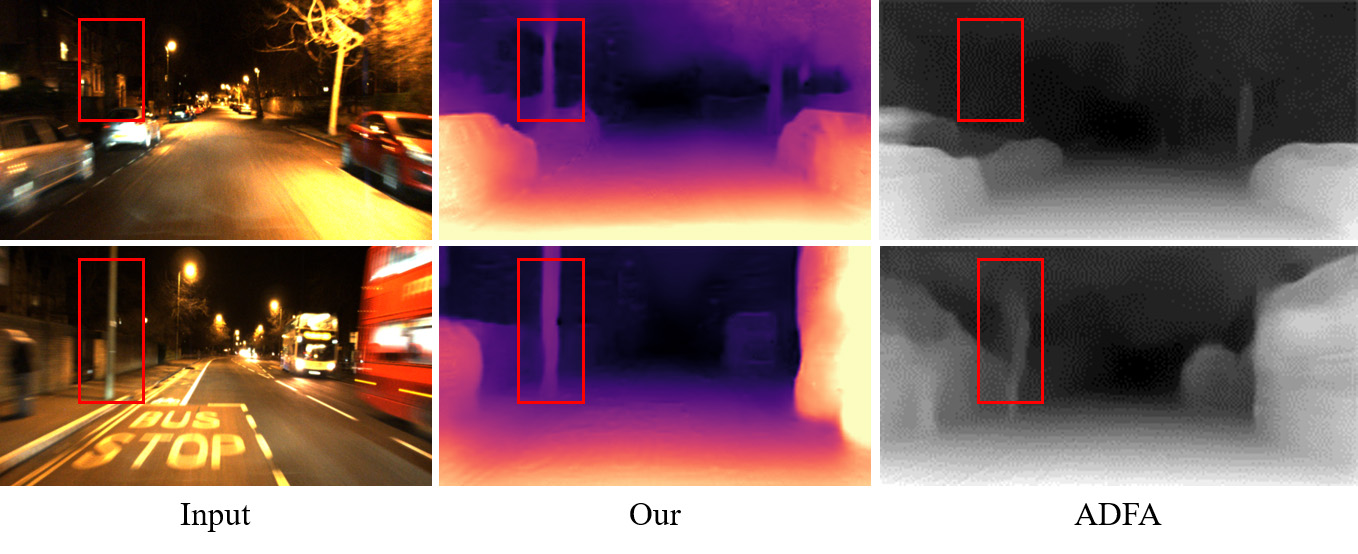}
    \end{center}
    \vspace{-0.2cm}
    \caption{Qualitative comparison on our method (middle) and ADFA \cite{adfa} (right). The images come from the Fig. 1 in ADFA.}
    \label{fig.compare_adfa}
    \vspace{-0.4cm}
\end{figure}

Furthermore, we conduct several validation experiments and report the results at the second part of RobotCar-Night. Models labeled with (Day) are trained on daytime dataset and directly tested on nighttime scenes. They score higher on the first four error metrics but worse on the last three accuracy metrics, indicating their weakness on prediction accuracy. See Supp for more analysis. Reg Only is trained with solely PBR regularization. It doesn't work well, since just constraining the distribution consistency with referenced depth maps is not enough to infer the depth of a specific image. This is also the reason we use photometric loss as primary constraint and PBR loss as regularization in our framework. 

% three validation experiments are conducted. Methods labeled with (Day) are trained on a daytime dataset. They obtain a good result on the first four metrics yet show inaccuracy on the last three. It denotes that the plan to employ models trained on another dataset is not advisable, due to the domain gap between two different datasets. Moreover, Reg Only is trained with solely PBR regularization and shows poor accuracy, which indicates that inferring the depth map of a specific image from a general distribution is unreasonable.

The qualitative results on RobotCar-Night and nuScenes-Night are reported in Figs. \ref{fig.compare_rc} and \ref{fig.compare_ns}, respectively. We compare our method with three SOTA approaches, including SC-SfMLearner \cite{bian}, MonoDepth2 \cite{monodepth2} and FM \cite{fm}. In general, the SOTAs fail to produce smooth depth maps and miss some details of objectives. By contrast, the proposed framework greatly alleviates the non-smoothness and produces higher quality depth outputs. In Fig. \ref{fig.compare_ns}, our model is still able to make a plausible guess on very dark scenes which are even challenging for human eye. It demonstrates the effectiveness of our method to regularize weird outputs in nighttime depth estimation.

More importantly, we compare with ADFA \cite{adfa} in qualitative results. It firstly focuses on nighttime depth estimation and leverages adversarial domain adaptation to address this problem. In the two samples of Fig. \ref{fig.compare_adfa}, ADFA produces blurry outputs and is unable to predict the accurate depth of the two objects framed by red boxes. In contrast, our results are clearer and more accurate. Compared to ADFA, the proposed method learns to predict depth directly from nighttime data instead of transferring knowledge learned from daytime scenarios. This enables models to better adapt to nighttime environments, thus achieves better performance.

\begin{figure*}
    \begin{center}
        \includegraphics[width=0.95\linewidth]{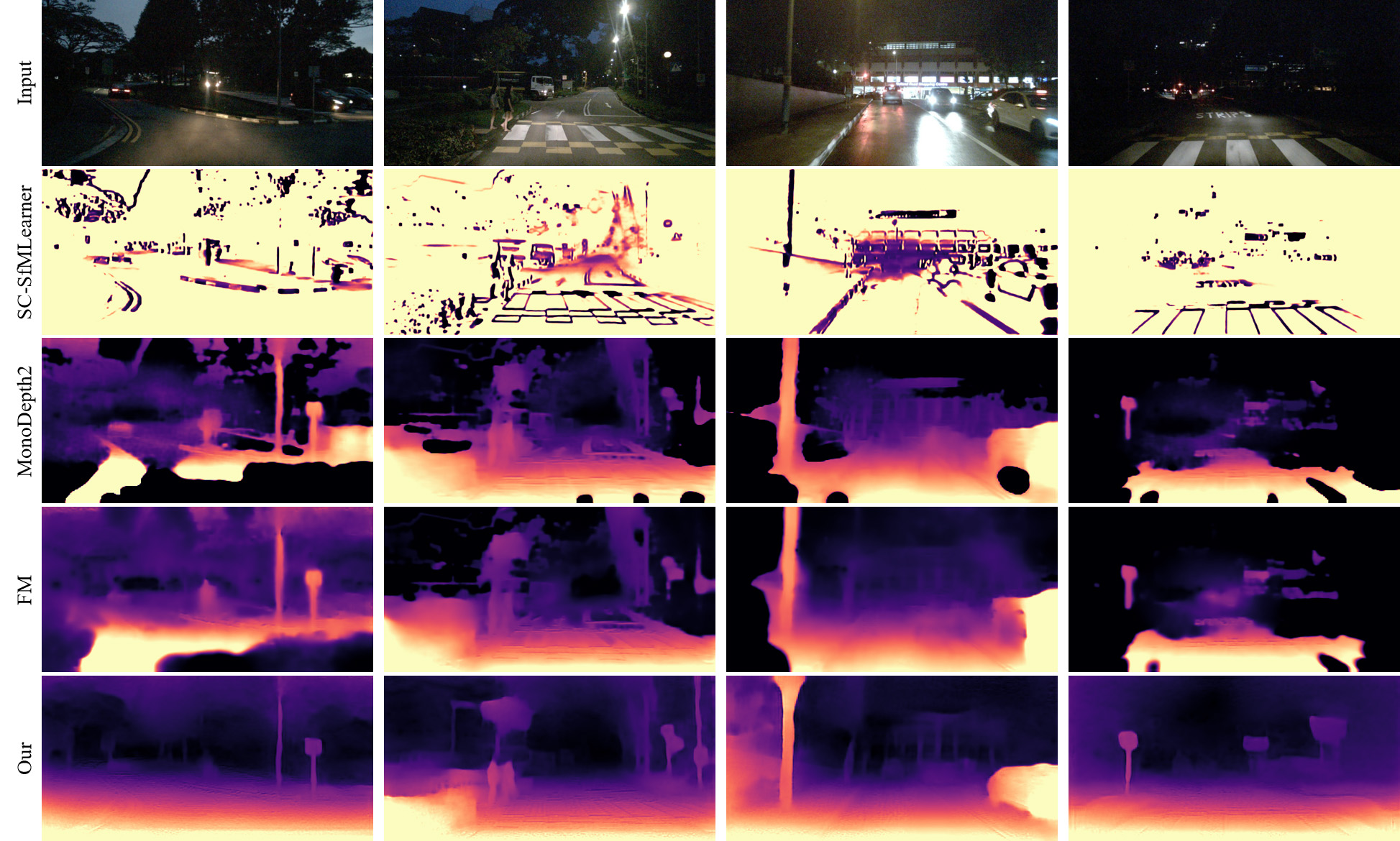}
    \end{center}
    \vspace{-0.2cm}
    \caption{Qualitative results on nuScenes-Night. This dataset is more challenging due to lower visibility and more complex traffic, while our method is still able to make plausible predictions.}
    \label{fig.compare_ns}
    \vspace{-0.25cm}
\end{figure*}

\begin{table}
    \resizebox{\linewidth}{!}{
        \footnotesize
        \renewcommand\arraystretch{1.2}
        \centering
        \begin{tabular}{c||c|c|c|c|c|c|c}
            \hline
            Method     & \cellcolor[RGB]{244,176,132}Abs Rel & \cellcolor[RGB]{244,176,132}Sq Rel & \cellcolor[RGB]{244,176,132}RMSE & \cellcolor[RGB]{244,176,132}RMSE log & \cellcolor[RGB]{155,194,230}$\delta_1$ & \cellcolor[RGB]{155,194,230}$\delta_2$ & \cellcolor[RGB]{155,194,230}$\delta_3$ \\ \hline
            \multicolumn{8}{c}{\cellcolor[RGB]{198,224,180}RobotCar-Night}   \\
            \underline{MonoDepth2}              & 0.400            & 7.451          & 6.642         & 0.443          & 0.744         & 0.892         & 0.928             \\
            \hline
            PBR Only                & \textbf{0.126}   & \textbf{0.539} & \textbf{2.953}& \textbf{0.168} & \textbf{0.865}& \textbf{0.970}& \textbf{0.990}    \\
            MCIE Only               & 0.377            & 6.735          & 6.530         & 0.425          & 0.728         & 0.884         & 0.931             \\
            SBM Only                & 0.348            & 5.389          & 5.896         & 0.400          & 0.742         & 0.898         & 0.935             \\
            \hline
            PBR + MCIE              & 0.122            & 0.528          & 2.914         & 0.165          & 0.875         & 0.969         & 0.989             \\
            Full without $I_p$      & 0.128            & 0.588          & 3.112         & 0.173          & 0.856         & 0.966         & 0.989             \\
            Full Method             & \textbf{0.121}   & \textbf{0.520} & \textbf{2.902}& \textbf{0.163} & \textbf{0.879}& \textbf{0.969}& \textbf{0.990}    \\
            \hline
            \multicolumn{8}{c}{\cellcolor[RGB]{198,224,180}nuScenes-Night}   \\
            \underline{MonoDepth2}              & 1.185            & 42.306         & 21.613        & 1.570         & 0.184         & 0.360         & 0.504            \\
            \hline
            PBR Only                & \textbf{0.325}   & \textbf{4.127} & \textbf{9.881}& \textbf{0.413}& \textbf{0.508}& \textbf{0.770}& \textbf{0.888}   \\
            MCIE Only               & 1.153            & 40.741         & 21.193        & 1.511         & 0.202         & 0.377         & 0.521            \\
            SBM Only                & 0.779            & 25.794         & 16.657        & 0.680         & 0.354         & 0.594         & 0.744            \\
            \hline
            PBR + MCIE              & 0.321            & 4.005          & 9.644         & 0.403         & 0.508         & \textbf{0.784}& \textbf{0.898}   \\
            Full without $I_p$      & 0.333            & 4.467          & 9.947         & 0.417         & \textbf{0.509}& 0.772         & 0.888            \\
            Full Method             & \textbf{0.315}   & \textbf{3.793} & \textbf{9.641}& \textbf{0.403}& 0.508         & 0.778         & 0.896             \\
            \hline
        \end{tabular}   
    }     
    \caption{Quantitative results of ablation study. Baseline method is \underline{underlined} and the best results in each part are in \textbf{bold}. $I_p$ denotes coordinates image in PBR. Full Method means all three (PBR, MCIE and SBM) components are enabled.}
    \label{tab.ablation}
    \vspace{-0.5cm}
\end{table}

\subsection{Ablation Study}
Here, we conduct a series of experiments to demonstrate the effectiveness of proposed components and report the results in Table \ref{tab.ablation}. Firstly, baseline method coupled with each individual component (PBR Only, MCIE Only and SBM Only) is tested. The results in the second part show improved performance, indicating the effectiveness of each component. Among them, PBR Only performs the best. It promotes the Sq Rel by $92.8\%$ and $90.2\%$ on RC and NS, respectively. Followed by SBM Only, then MCIE Only, the former separately obtains an improvement of $27.7\%$ and $39.0\%$ while the later $9.6\%$ and $3.7\%$ on these two datasets.

% experiments in which the baseline method is combined with each individual component are implemented, including PBR Only, MCIE Only and SBM Only. PBR module achieves the most improvement compared with baseline and followed by SBM. The results confirm that the method to bring in prior distribution knowledge for regularizing model training is advisable. Besides, MCIE and SBM manifest their ability to deal with nighttime scenarios by presenting better performance. The MCIE module shows better result on RC, because the less noise in it allows MCIE to run with less limitation.

Next, we further evaluate the framework by gradually enabling each component. The results are reported at PBR Only, PBR + MCIE and Full Method. In summary, the performance is improved as more components are enabled. On RC, $3.5\%$ improvements of Sq Rel are achieved by Full Method when compared to PBR Only and $1.5\%$ in comparison with PBR + MCIE. As for NS, the proportion is $8.1\%$ and $5.3\%$, respectively.

Also, the coordinates image $I_p$ is tested through Full without $I_p$ and Full Method. Compared with the later, the Sq Rel metric of the former drops by $11.6\%$ on RC and $15.1\%$ on NS, respectively. This validates the association between image coordinates and depth distribution.

% Moreover, further experiments are designed to verify the impact of each component on the entire framework. We gradually enable each component and report the results in PBR Only, PBR + MCIE and Full Method. In general, as each component is used, the performance of the framework gradually improves. MCIE shows advantages on RMSE metric, which indicates that it helps to reduce the overall prediction error. SBM achieves the greatest improvement on Sq Rel metric and denotes that it has a positive effect on reducing large errors.

% Furthermore, we extend experiments to validate the effectiveness of coordinates image $I_p$ in PBR. The results are reported in Full without $I_p$. Compared with Full Method, its performance drops obviously, especially on Sq Rel metric, on both two datasets. $I_p$ is introduced to improve the sensitivity of PBR to depth distribution which is related to pixel position and experiments confirm it by showing better results.

%------------------------------------------------------------------------
\section{Conclusion}
In this paper, we propose a novel framework with three improvements to effectively address the problem of self-supervised nighttime depth estimation. Priors-Based Regularization leverages prior distribution from referenced depth maps to regularize model training; Mapping-Consistent Image Enhancement module enhances image brightness and contrast while maintaining brightness consistency to deal with the low visibility in the dark; Statistics-Based Mask flexibly removes pixels within textureless regions using dynamic statistics to mitigate depth ambiguity. Benefits from these improvements,  our method significantly outperforms current SOTA methods and greatly alleviate the weird outputs in nighttime depth estimation.

%------------------------------------------------------------------------
\section{Acknowledgement}
The authors would like to thank the editor and the anonymous reviewers for their critical and constructive comments and suggestions. This work was supported by the National Science Fund of China under Grant No. U1713208, 62072242 and Postdoctoral Innovative Talent Support Program of China under Grant BX20200168, 2020M681608.

%------------------------------------------------------------------------
{\small
    \bibliographystyle{ieee_fullname}
    \typeout{}
    \bibliography{references}
}

%------------------------------------------------------------------------
% Supplementary Meterial
\clearpage
\appendix

\section{Dataset Construction}

Dataset for self-supervised monocular depth training in nighttime is under-explored. To make up this lack, we build two nighttime datasets, named RobotCar-Night (RC-N) and nuScenes-Night (NS-N). The two datasets consist of many video clips from Oxford RobotCar \cite{RobotCarDatasetIJRR} and nuScenes \cite{nuscenes2019}, along with carefully generated ground truth using the official toolbox\footnote{Oxford RobotCar: \href{https://github.com/ori-mrg/robotcar-dataset-sdk}{https://github.com/ori-mrg/robotcar-dataset-sdk}, nuScenes: \href{https://github.com/nutonomy/nuscenes-devkit}{https://github.com/nutonomy/nuscenes-devkit}}.

\begin{figure}[H]
    \begin{center}
        \includegraphics[width=0.95\linewidth]{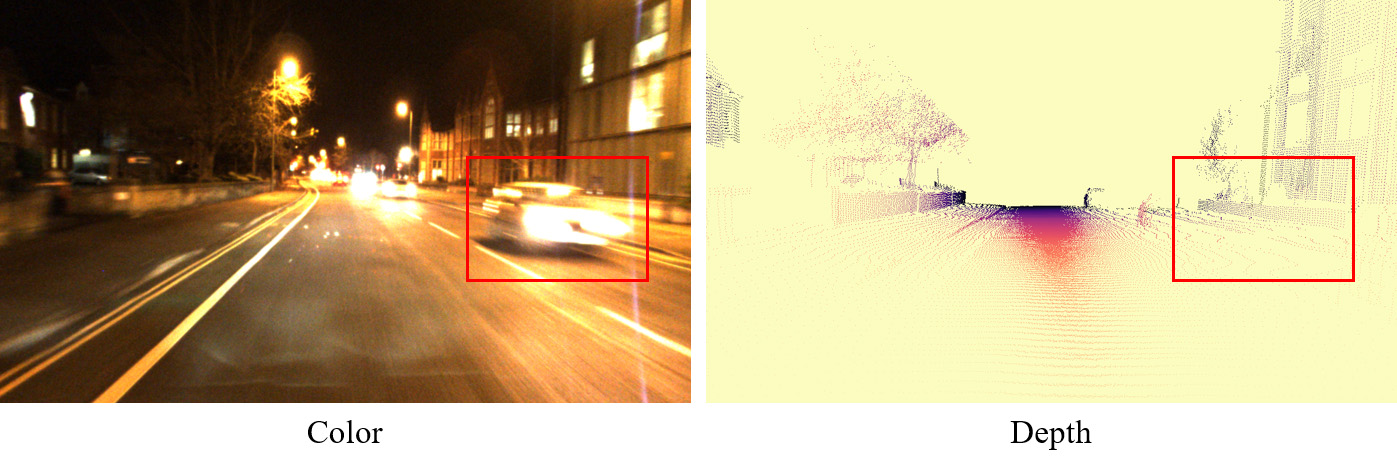}
    \end{center}
    \vspace{-0.3cm}
    \caption{Sample from Oxford RobotCar containing moving objects. The \textcolor{red}{red} box indicates a wrong construction of depth.}
    \label{fig.robotcar}
\end{figure}

In RobotCar, the number of LiDAR points in one frame is relatively small, so multiple frames are combined to generate the ground truth depth using official scripts. This process is based on Structure-from-Motion (SFM), therefore moving objects lead to wrong outputs. For example, a generated depth map is visualized in Fig. \ref{fig.robotcar}. It shows an obvious mistake on the moving car framed by a red box. To tackle this problem, we manually select scenes without moving object and carefully pick up many high-quality outputs among them. This approach is different from the previous work \cite{adfa}, in which random sampling is used to choose test samples.

\noindent\textbf{Remark}. In main text, there are several samples containing moving objects in Fig. 5. These samples are from the same video sequences as RC-N but neither included in test nor training split. This is also the case for the last sample in Fig. \ref{fig.robotcar}.

By contrast, the LiDAR data in nuScenes contain more than 3,000 valid points in one frame and covers a wide range of depth values. Thus, data from single frame is used to prepare the ground truth depth maps and random sampling is applied to form the final test split.

Furthermore, some video clips containing daytime scenarios are selected from Oxford RobotCar and nuScenes to separately build RobotCar-Day (RC-D) and nuScenes-Day (NS-D), which are used to generate referenced depth maps.

%--------------------------------------------------------------------------
\section{Parameter Setting}
Here, we discuss the parameter setting about $\sigma$ in MCIE and $\epsilon$ in SBM. Images captured in low light environments are usually noisy, thus a smaller $\sigma$ should be set to avoid an excessive enhancement on noise. In darker scenarios, more textureless pixels need to be masked out. Therefore, $\epsilon$ should be set to a bigger value. In our experiment, $(\sigma, \epsilon)$ is set to $(0.008, 10)$ and $(0.004, 20)$ on RC-N and NS-N dataset, respectively. This can be a empirical reference to set these two parameters.

\begin{figure}
    \begin{center}
        \includegraphics[width=0.98\linewidth]{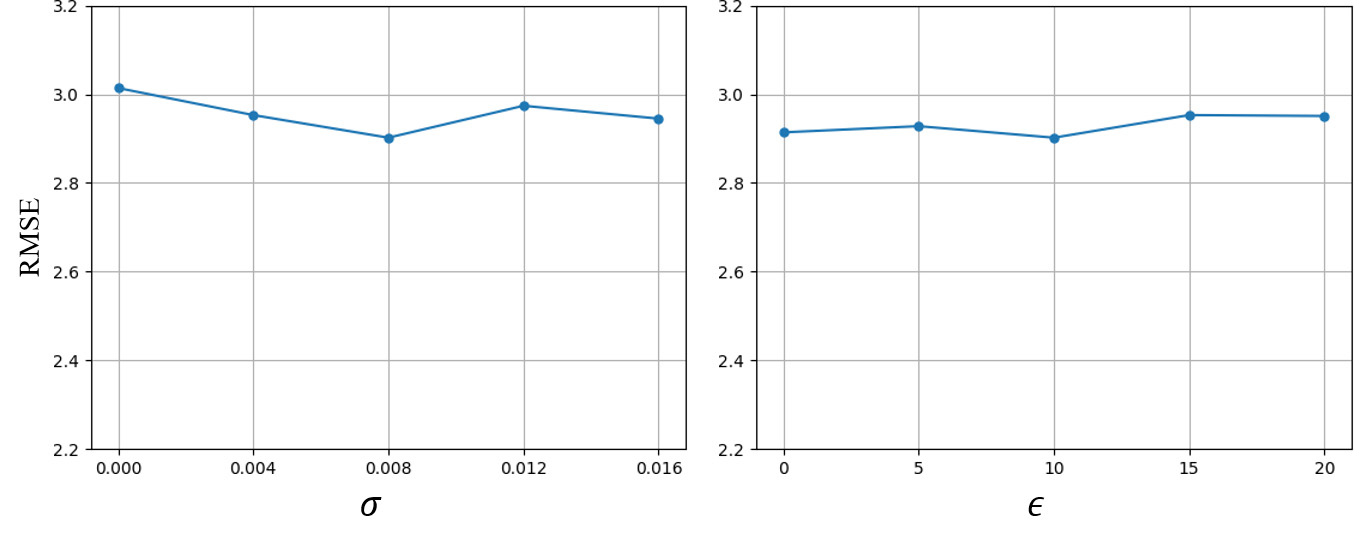}
    \end{center}
    \vspace{-0.3cm}
    \caption{The left and right chart separately show the effect of $\sigma$ and $\epsilon$. The y axis is RMSE error and the x axis denotes different values of these two parameters. }
    \label{fig.param}
\end{figure}

To explore the effect of these two parameters, we conduct series comparison tests on RC-N and report the RMSE error in Fig. \ref{fig.param}. The variables in the left and right chart are $\sigma$ and $\epsilon$, respectively. Zero indicates the corresponding module is not enabled. Overall, these two parameters impact little to the framework which performs the best when $\sigma=0.008, \epsilon=10$.

Generally speaking, $[0.002, 0.01]$ and $[10, 20]$ are proper ranges for $\sigma$ and $\epsilon$, respectively. Besides, comparison tests can help to choose the best parameter setting.

\begin{figure*}
    \begin{center}
        \includegraphics[width=0.9\linewidth]{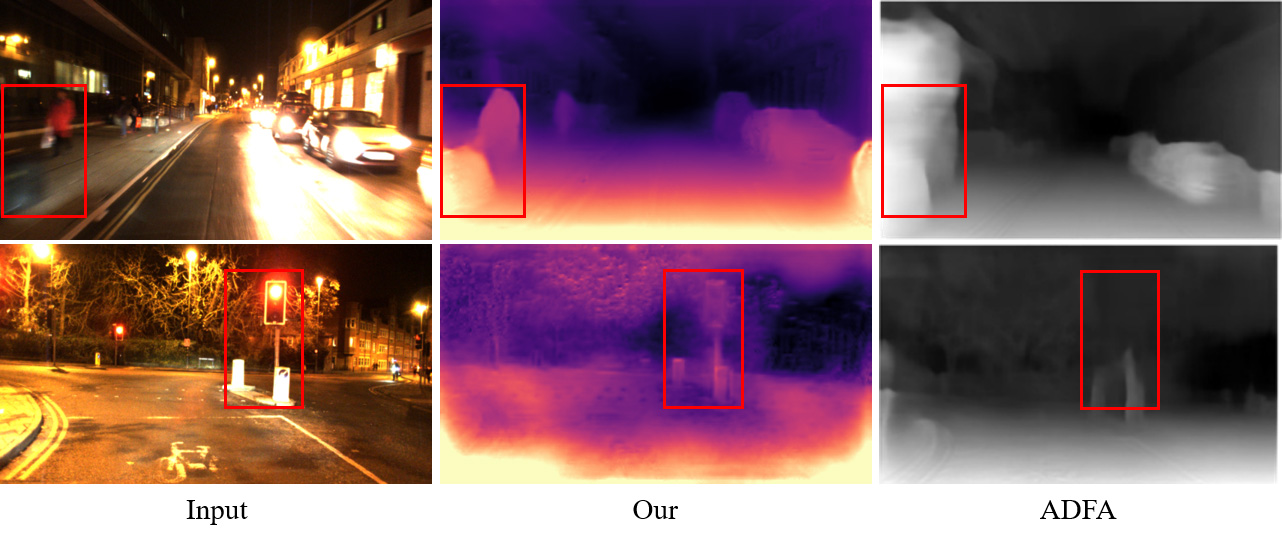}
    \end{center}
    \vspace{-0.4cm}
    \caption{Qualitative comparison between our method and ADFA \cite{adfa} on two samples containing saturated and blurred regions. These two images come from the Fig. 4 of ADFA.}
    \label{fig.compare_failure}
\end{figure*}

%-------------------------------------------------------------------------
\section{Selection of Referenced Scene}
In our framework, the reference depth maps are generated by a depth estimation network $\Phi'_d$ trained on RC-D and NS-D in a self-supervised manner. They provide prior knowledge about depth distributions and are unpaired with nighttime scenarios. Generally speaking, depth maps in various driving scenes can be used as references, since they share similar depth distributions. To explore the effect of different reference scenarios, we train the framework in two referenced scenarios and report quantitative results in Tab. \ref{tab.compare_scenes}. The method \emph{Our (RobotCar-Day)} achieves a slightly worse but similar performance compared to \emph{Our (nuScenes-Day)} and significantly outperforms other SOTA methods. This illustrates that depth maps from other scenarios are also able to regularize training.
\begin{table}[H]
    \centering
    \resizebox{0.98\linewidth}{!}{
        \footnotesize
        \renewcommand\arraystretch{1.2}
        \begin{tabular}{c||c|c|c|c|c|c|c}
            \hline
            Method     & \cellcolor[RGB]{244,176,132}Abs Rel & \cellcolor[RGB]{244,176,132}Sq Rel & \cellcolor[RGB]{244,176,132}RMSE & \cellcolor[RGB]{244,176,132}RMSE log & \cellcolor[RGB]{155,194,230}$\delta_1$ & \cellcolor[RGB]{155,194,230}$\delta_2$ & \cellcolor[RGB]{155,194,230}$\delta_3$ \\ \hline
            MonoDepth2 \cite{monodepth2}  & 1.1848            & 42.3059           & 21.6129           & 1.5699            & 0.1842            & 0.3598            & 0.5044           \\ 
            SfMLearner \cite{zhou_method}   & 0.6004            & 8.6346            & 15.4351           & 0.7522            & 0.2145            & 0.4166            & 0.5961            \\
            SC-SfMLearner \cite{bian}   & 1.0508            & 30.5865           & 19.6004           & 0.8854            & 0.1823            & 0.3673            & 0.5422            \\
            PackNet \cite{3dpack}         & 1.5675            & 61.5101           & 25.8318           & 1.3717            & 0.1387            & 0.2980            & 0.4313            \\
            FM \cite{fm}             & 1.1383            & 41.6166           & 20.8481           & 1.1483            & 0.2376            & 0.4252            & 0.5650            \\
            \hline
            Our (nuScenes-Day)        & \textbf{0.3150}  & \textbf{3.7926}   & \textbf{9.6408}   & \textbf{0.4026}   & \underline{0.5081}            & \textbf{0.7776}   & \textbf{0.8959} \\
            Our (RobotCar-Day)        & \underline{0.3285}           & \underline{4.3069}        & \underline{10.2651} & \underline{0.4197}            & \textbf{0.5142}   & \underline{0.7642}            & \underline{0.8813} \\
            \hline
        \end{tabular} 
    }
    \caption{Quantitative results on nuScenes-Night, using depth maps from nuScenes-Day and RobotCar-Day as references, respectively.}
    \label{tab.compare_scenes}
\end{table}
%-------------------------------------------------------------------------

\section{Comparison with ADFA in Challenging Cases}

\begin{figure}
    \centering
    \includegraphics[width=0.95\linewidth]{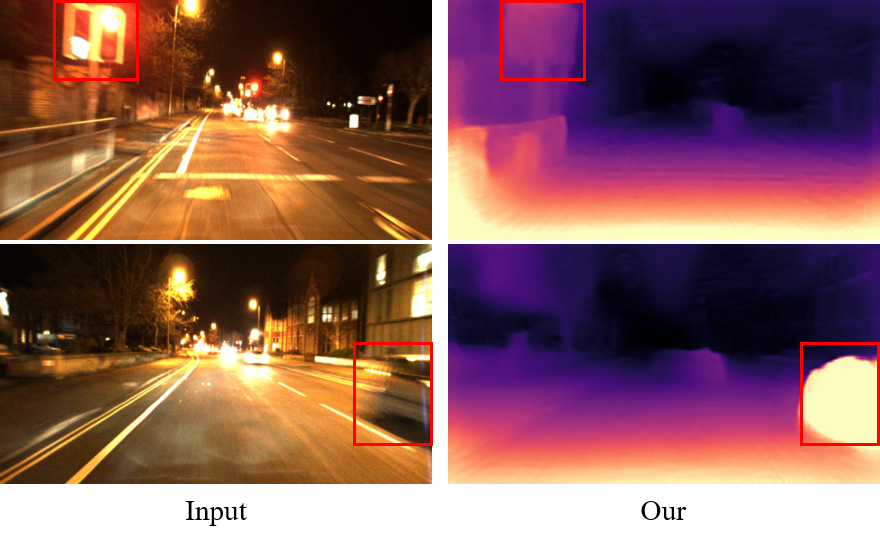}
    \caption{Qualitative results of our method on two samples containing saturated and blurred regions. These two images are from RobotCar-Night.}
    \label{fig.challenging_cases}
\end{figure}

ADFA \cite{adfa} claims three challenging cases that lead to its failure, including nighttime images with very low-illumination conditions, blurred image regions and saturated regions (bright light spots). Here, we further compare our method with ADFA in these three cases.

\noindent\textbf{Blurred and Saturated Image Regions}. Fig. \ref{fig.compare_failure} shows a comparison on two image samples containing saturated and blurred regions. Compared with ADFA, our method achieves better performance on presenting the shape of objects. Furthermore, two similar samples are shown in Fig. \ref{fig.challenging_cases} to further illustrate the advantages of our method.

\begin{figure}
    \centering
    \includegraphics[width=0.95\linewidth]{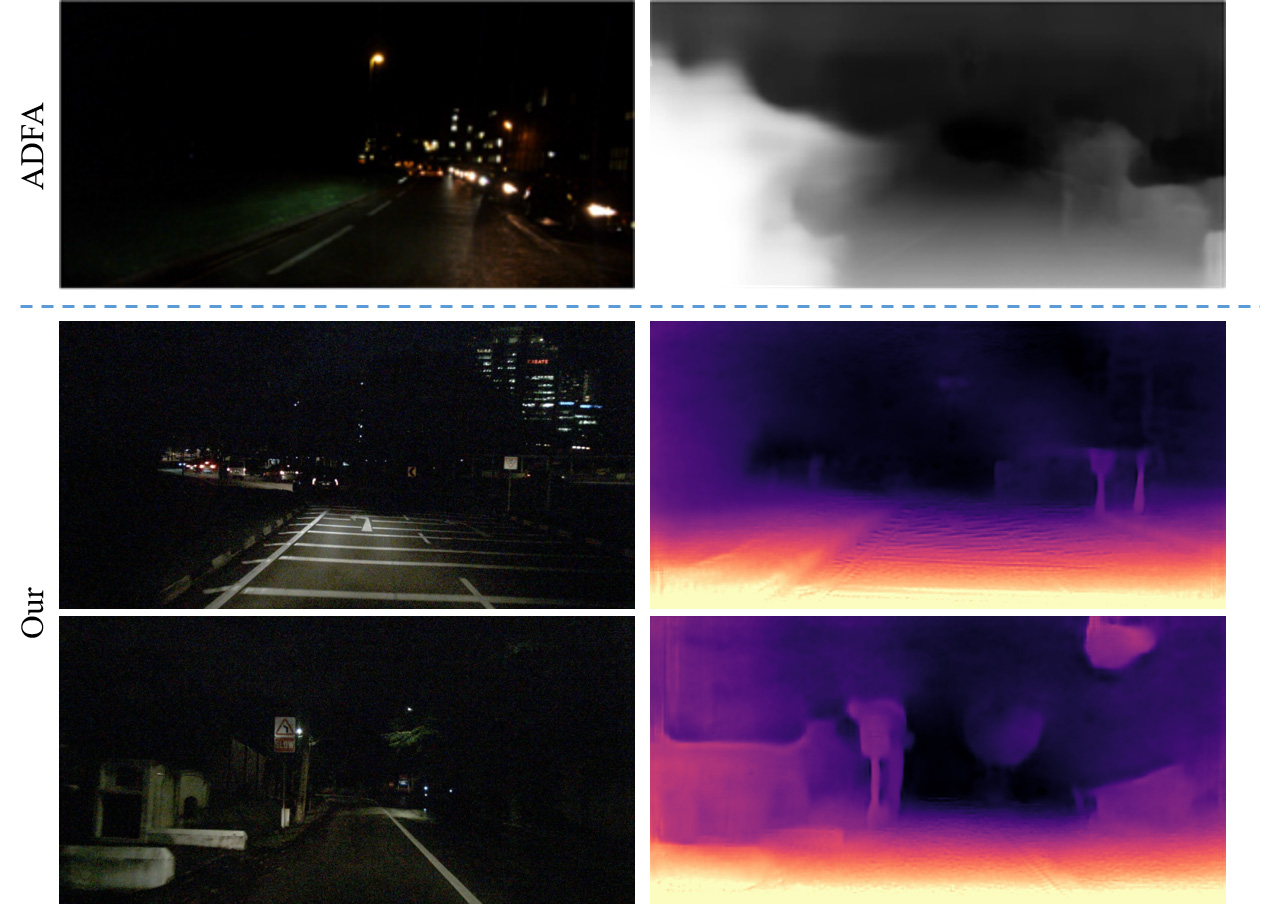}
    \caption{Qualitative results on very low-light images, where the top one and last two depth maps are generated by ADFA \cite{adfa} and our method, respectively. The first image comes from the Fig. 4 of ADFA \cite{adfa} and the last two images are from nuScenes-Night.}
    \label{fig.low_light}
\end{figure}

\noindent\textbf{Images with very Low-Illumination Conditions}. Very low-light images are a huge challenge for self-supervised depth estimation. Fig. \ref{fig.low_light} shows three samples in very low illuminated environments, where the top one and last two are generated by ADFA and our method, respectively. On the first sample, ADFA produces a blurry and inaccurate depth map. In contrast, our method is still able to make a plausible prediction on the second sample. The last sample is captured in a very dark environment. Our method makes a coarse estimation on some objectives but fails to depict the depth of entire scene.

%-------------------------------------------------------------------------
\section{More Qualitative Result}
Here, we show more qualitative results on RobotCar-Night and nuScenes-Night datasets in Fig. \ref{fig.compare_rc} and Fig. \ref{fig.compare_ns}, respectively. Five SOTA methods are evaluated for comparison, including SfMLearner \cite{zhou_method}, SC-SfMLearner \cite{bian}, PackNet \cite{3dpack}, MonoDepth2 \cite{monodepth2} and FM \cite{fm}.

\begin{figure*}
    \begin{center}
        \includegraphics[width=0.95\linewidth]{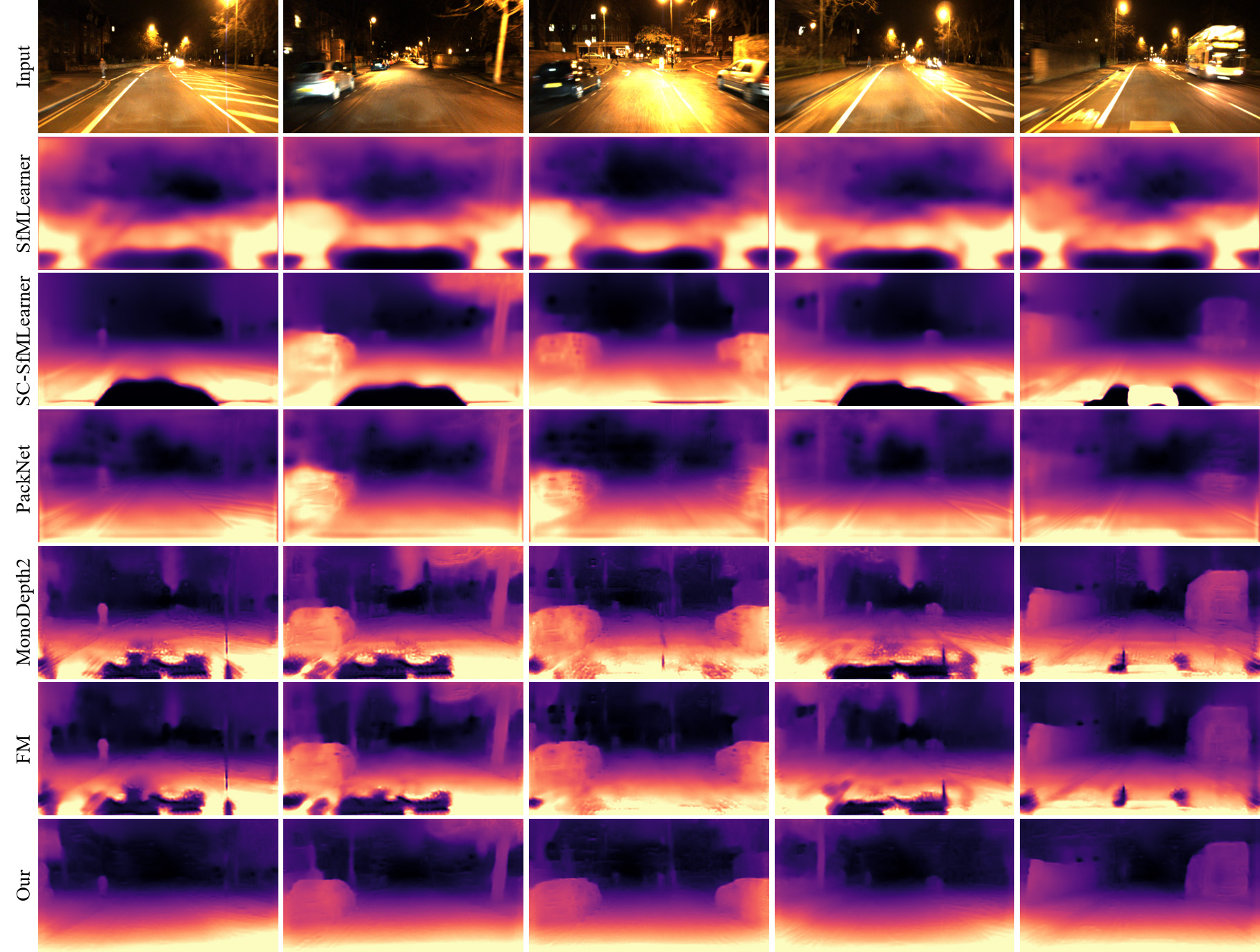}
    \end{center}
    \vspace{-0.25cm}
    \caption{Qualitative comparison on RobotCar-Night.}
    \label{fig.compare_rc}
\end{figure*}

\begin{figure*}
    \begin{center}
        \includegraphics[width=0.95\linewidth]{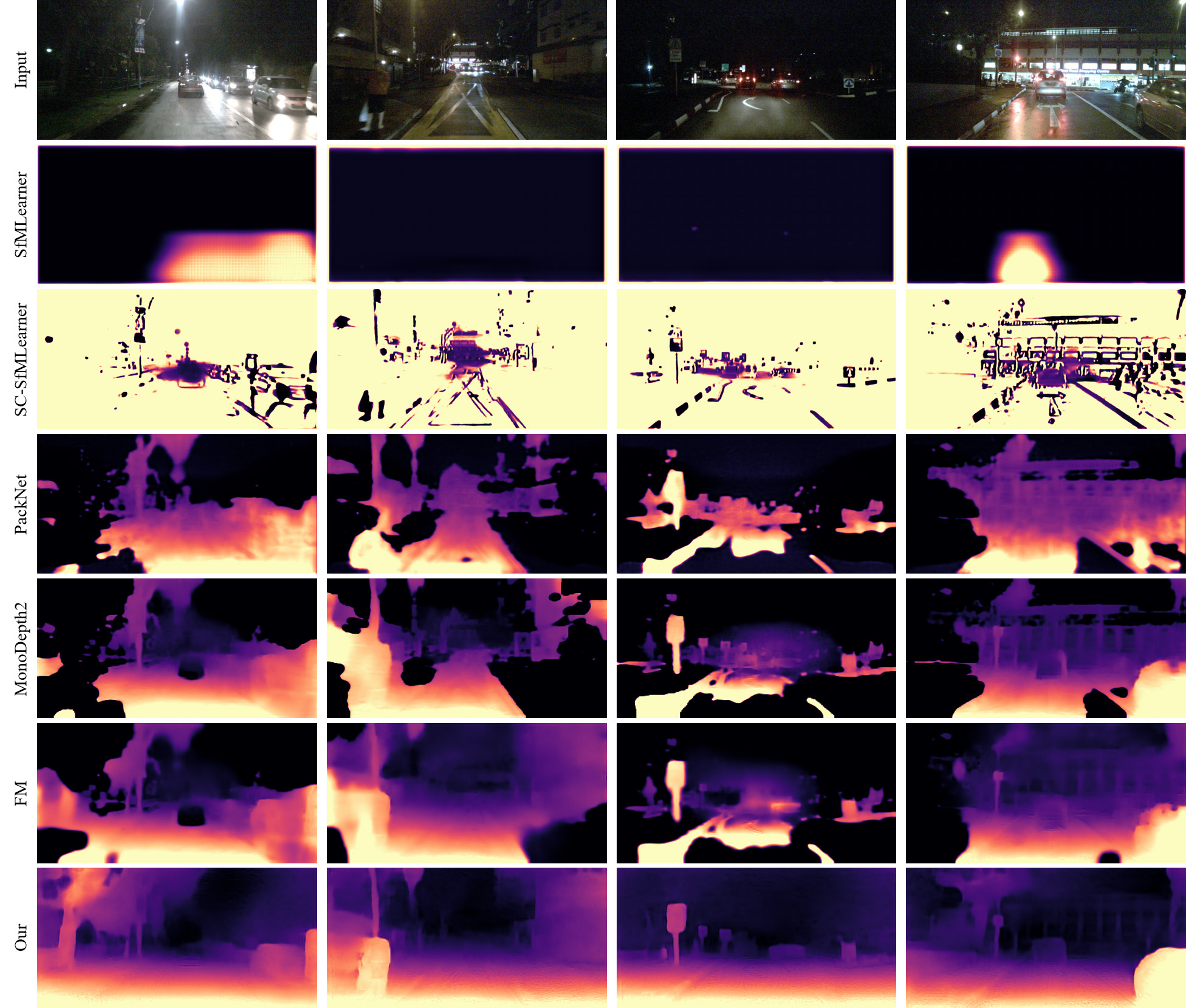}
    \end{center}
    \vspace{-0.25cm}
    \caption{Qualitative comparison on nuScenes-Night.}
    \label{fig.compare_ns}
\end{figure*}

%-------------------------------------------------------------------------
\section{Evaluation Metrics}
There are seven standard metrics are used for evaluation, including Abs Rel, Sq Rel, RMSE, RMSE log, $\delta_1$, $\delta_2$ and $\delta_3$, which are presented by
\begin{equation}\label{equ.metrics}
    \begin{split}
        \text{Abs Rel}=\frac{1}{|D|}\sum\nolimits_{d^*\in D}{|d^*-d|/d^*},\\
        \text{Sq Rel}=\frac{1}{|D|}\sum\nolimits_{d^*\in D}{\|d^*-d\|^2/d^*}, \\
        \text{RMSE}=\sqrt{\frac{1}{|D|}\sum\nolimits_{d^*\in D}{\|d^*-d\|^2}},\\
        \text{RMSE log}=\sqrt{\frac{1}{|D|}\sum\nolimits_{d^*\in D}{\|logd^*-logd\|^2}}, \\
        \delta_i=\frac{1}{|D|}|\{d^*\in D \max (\frac{d^*}{d},\frac{d}{d^*})<1.25^i \}|,
    \end{split}
\end{equation}
where $d$ and $d*$ separately denotes predicted and ground truth depth maps, $D$ indicates a set of valid ground truth depth values in one image, $|.|$ returns the number of elements in the input set.

%-------------------------------------------------------------------------
\section{More discussion on experiment results}
\begin{figure*}
    \centering
    \includegraphics[width=0.9\linewidth]{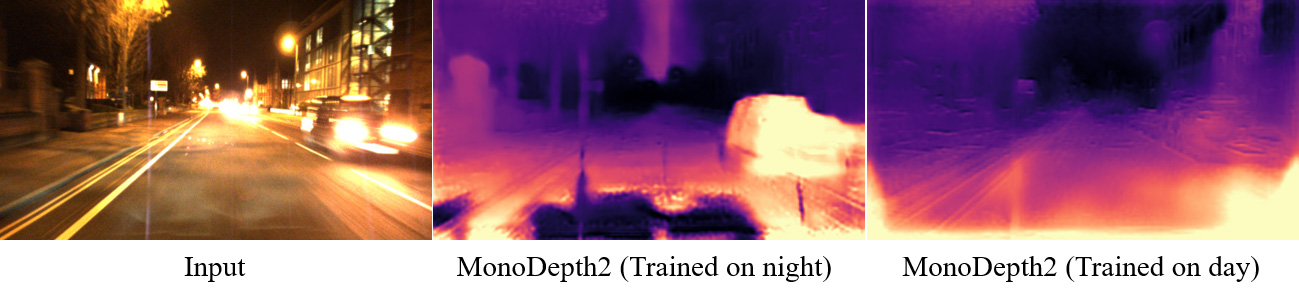}
    \caption{Qualitative comparison between MonoDepth2 (middle) and MonoDepth2 (Day) (Right).}
    \vspace{-0.25cm}
    \label{fig.compare_day}
\end{figure*}

\textbf{More Analysis on Experiments.} In Table. 1 of main text, we show the evaluation results on RC-N. One may notice that, \emph{MonoDepth2 (Day)} and \emph{FM (Day)} achieve better results on the first four error metrics yet worse on the last three accuracy ones than their counterparts. Here, we present a possible explanation on this phenomenon. Fig. \ref{fig.compare_day} shows two depth maps generated by MonoDepth2 (abbr. MD2) and MonoDepth2 (Day), respectively. The former produces more detailed results but with big holes while the later generates blurry outputs without holes. The big holes indicate a very large depth value and differ greatly from the Ground Truth, thus MD2 gets higher average errors on Abs\_Rel, Sq\_Rel, RMSE and RMSE\_log. Conversely, $\sigma_{1,2,3}$ denote the percentage of pixels below a certain threshold, therefore MD2 outperforms MD2 (Day) by more accurate predictions within non-hole areas.

\textbf{Mixed Data Training.} We train MD2 and FM with mixed daytime and nighttime data from nuScenes and report the results (\emph{MonoDepth2 (Mix)} and \emph{FM (Mix)}) in Table. \ref{tab.mix}. These two methods achieve better performance than their baselines but still keep a large gap to \emph{Our}.

\begin{table}[H]
    \begin{center}
        \resizebox{0.98\linewidth}{!}{
        \footnotesize
        \renewcommand\arraystretch{1.2}
        \begin{tabular}{c||c|c|c|c|c|c|c}
            \hline
            Method     & \cellcolor[RGB]{251,229,214}Abs Rel & \cellcolor[RGB]{251,229,214}Sq Rel & \cellcolor[RGB]{251,229,214}RMSE & \cellcolor[RGB]{251,229,214}RMSE log & \cellcolor[RGB]{222,235,247}$\delta_1$ & \cellcolor[RGB]{222,235,247}$\delta_2$ & \cellcolor[RGB]{222,235,247}$\delta_3$ \\ \hline
            \underline{MonoDepth2}    & 1.185   & 42.306         & 21.613    & 1.570    & 0.184     & 0.360     & 0.504            \\
            \underline{FM}            & 1.138   & 41.617        & 20.848   & 1.148   & 0.238    & 0.425    & 0.5650            \\
            \hline
            MonoDepth2 (Mix)          & 1.070   & 38.336        & 20.117     & 1.191    & 0.269     & 0.451     & 0.586   \\
            FM (Mix)                  & 0.956   & 34.052         & 18.794    & 0.798    & 0.305     & 0.507     & 0.652            \\
            \hline
            Our             & \textbf{0.315}   & \textbf{3.793}   & \textbf{9.641}   & \textbf{0.403}   & \textbf{0.508}   & \textbf{0.778}   & \textbf{0.896}\\
            \hline
        \end{tabular}
    }
    \caption{Quantitative results on mixed daytime and nighttime data of nuScenes. Baseline methods are \underline{underlined}.}
    \label{tab.mix}
    \end{center}
\end{table}

%-------------------------------------------------------------------------

\end{document}